\documentclass[runningheads]{llncs}
\usepackage[T1]{fontenc}
\usepackage{graphicx}
\usepackage[noend]{algpseudocode}
\usepackage{algorithm}
\usepackage{booktabs}
\usepackage{makecell}
\usepackage{wrapfig}
\usepackage{float}
\usepackage{hyperref}
\usepackage{todonotes}
\usepackage[binary-units=true,detect-weight=true]{siunitx}
\usepackage[rightcaption]{sidecap}
\sidecaptionvpos{figure}{r}

\usepackage{xcolor}
\usepackage{listings}
\usepackage{amssymb}
\definecolor{keywordcolor}{rgb}{0.7, 0.1, 0.1}
\definecolor{tacticcolor}{rgb}{0.0, 0.1, 0.6}
\definecolor{commentcolor}{rgb}{0.4, 0.4, 0.4}
\definecolor{symbolcolor}{rgb}{0.0, 0.1, 0.6}
\definecolor{sortcolor}{rgb}{0.1, 0.5, 0.1}
\definecolor{attributecolor}{rgb}{0.7, 0.1, 0.1}

\begin{document}
\title{Machine-Learned Premise Selection for Lean\thanks{
The results were supported by the Hoskinson Center for Formal Mathematics
(BP, RFM, EA), the Kościuszko Foundation (BP), and the Principal's Career Development
Scholarship of the University of Edinburgh (RFM).}}
\author{
Bartosz Piotrowski\inst{1} \and
Ramon Fernández Mir\inst{2} \and
Edward Ayers\inst{3}
}

\authorrunning{Piotrowski, Fernández Mir, Ayers}
\institute{
University of Warsaw and Czech Technical University \\
\email{bartoszpiotrowski@post.pl} \and
University of Edinburgh \and
Carnegie Mellon University
}
\maketitle
\begin{abstract}
We introduce a machine-learning-based tool for the Lean proof assistant that suggests relevant premises for theorems being proved by a user.
The design principles for the tool are (1) tight integration with the proof assistant, (2) ease of use and installation, (3) a lightweight and fast approach.
For this purpose, we designed a custom version of the random forest model, trained in an online fashion.
It is implemented directly in Lean, which was possible thanks to the rich and efficient metaprogramming features of Lean 4.
The random forest is trained on data extracted from \textsf{mathlib} -- Lean's mathematics library.
We experiment with various options for producing training features and labels.
The advice from a trained model is accessible to the user via the \lstinline{suggest_premises} tactic which can be called in an editor while constructing a proof interactively.
\end{abstract}
\section{Introduction}
\label{sec:intro}

Formalizing mathematics in proof assistants is an ambitious and hard undertaking.
One of the major challenges in constructing formal proofs of theorems depending on multiple other results is the prerequisite of having a good familiarity with the structure and contents of the library.
Tools for helping users search through formal libraries are currently limited.

In the case of the Lean proof assistant \cite{lean4}, users may look for relevant lemmas in its formal library, \textsf{mathlib} \cite{mathlib}, either by (1) using general textual search tools and keywords, (2) browsing the related source files manually, (3) using \textsf{mathlib}'s \lstinline{suggest} or \lstinline{library_search} tactics.

Approaches (1) and (2) are often slow and tedious.
The limitation of approach (3) is the fact that \lstinline{suggest} or \lstinline{library_search} propose lemmas that strictly match the goal at the current proof state.
This is often very useful, but it also means that these tactics often fail to direct the user to relevant lemmas that do not match the current goal exactly.
They may also suggest too many trivial lemmas if the goal is simple.

The aim of this project is to make progress towards improving the situation of a Lean user looking for relevant lemmas while building proofs.
We develop a new tool that efficiently computes a ranking of potentially useful lemmas selected by a machine learning (ML) model trained on data extracted from \textsf{mathlib}.
This ranking can be accessed and used interactively via the \lstinline{suggest_premises} tactic.

The project described here belongs to the already quite broad body of work dealing with the problem of fact selection for theorem proving \cite{premsel-flyspeck,premsel-alama,atpboost,deepmath}.
This problem, commonly referred to as the \textit{premise selection} problem,
is a crucial when performing automated reasoning in large formal libraries, both in the context of \textit{automated} (ATP) and \textit{interactive} (ITP) theorem proving, regardless of the underlying logical calculus.
Most of the existing work on premise selection focuses on the ATP context.
Our main contribution is the development of a premise selection tool that is practically usable in a proof assistant, tightly integrated with it, lightweight, extendable, and equipped with a convenient interface.

\section{Dataset Collection}
\label{sec:data}

A crucial requirement of a useful ML model is a high-quality dataset of training examples.
It should represent the learning task well and be suitable for the ML architecture being applied.

In this work, we use simple ML architectures that cannot process raw theorem statements and require \textit{featurization} as a preprocessing step.
The features need to be meaningful yet simple so that the model can use them appropriately.
Our approach is described in Section \ref{sec:features}.
The notion of \textit{relevant premise} may be understood differently depending on the context.
In Section \ref{sec:filters}, we describe the different specifications of this notion that we used in our experiments.

The tool developed in this work is implemented and meant to be used in Lean~4 together with \textsf{mathlib}~4. 
However, since, at the time of writing, Lean~4’s version of the library is still being ported from Lean 3, we use \textsf{mathlib3port}\footnote{\url{https://github.com/leanprover-community/mathlib3port} (commit \texttt{f4e5dfe})} as our main data source.

\subsection{Features}
\label{sec:features}

The features, similar to those used in \cite{enigma,atpboost}, consist of the symbols used in the theorem statement with different degrees of structure.
In particular, three types of features are used: \texttt{names}, \texttt{bigrams} and \texttt{trigrams}.
As an illustration, take this theorem about groups with zero:

\begin{lstlisting}
theorem div_ne_zero (ha : a ≠ 0) (hb : b ≠ 0) : a / b ≠ 0 := ...
\end{lstlisting}

The most basic form of featurization is the bag-of-words model, where we simply collect all the \texttt{names} (and numerical constants) involved in the theorem.
Following this definition naively, we would obtain \lstinline{≠}, \lstinline{0} and \lstinline{/}.
The reality is a bit more complicated as there are names in the expression that are hidden by the notation, e.g. \lstinline{OfNat.ofNat}.

Some of the \texttt{names} in \lstinline{div_ne_zero} are listed below. The ones starting with \lstinline{T} appear in the conclusion, while the ones starting with \lstinline{H} appear in the hypotheses.

\begin{lstlisting}
H:OfNat.ofNat H:MonoidWithZero.toZero H:0 H:Ne T:HDiv.hDiv T:0 T:Ne ...
\end{lstlisting}

It would be desirable, however, to keep track of which symbols appear next to each other in the syntactic trees of the theorem hypotheses and its statement.
Thus, we extract \texttt{bigrams} that are formed by the head symbol and each of its arguments (separated by \texttt{/} below).

\begin{lstlisting}
H:Ne/OfNat.ofNat H:OfNat.ofNat/0 T:OfNat.ofNat/0 T:Ne/OfNat.ofNat ...
\end{lstlisting}

Similarly, we also consider \texttt{trigrams}, taking all paths of length 3 from the syntactic tree of the expression.

\begin{lstlisting}
H:Ne/OfNat.ofNat/0 H:Ne/OfNat.ofNat/Zero.toOfNat0 ...
\end{lstlisting}

\subsection{Relevant Premises}
\label{sec:filters}

To obtain the list of all the premises used in the proof term it suffices to traverse the Lean expression and keep track of all the constants whose type is a proposition.
For instance, the raw list of premises that appear in the proof of \lstinline{div_ne_zero} is:
\begin{lstlisting}
GroupWithZero.noZeroDivisors mul_ne_zero inv_ne_zero Eq.refl div_eq_mul_inv
\end{lstlisting}

For more complicated examples, this approach results in a large number of premises including lemmas used \textit{implicitly} by tactics (for instance, those picked by the `simplify' tactic \lstinline{simp}), or simple facts that a user would rarely write explicitly.
Three different filters are applied to mitigate this issue: \texttt{all}, \texttt{source}, and \texttt{math}.
They are described below and their overall effect is shown in Table \ref{tab:filters}.

\begin{table}[b!]
    \caption{\label{tab:filters}
    Filters' statistics.
    An example is a theorem with a non-empty list of premises.}
    \centering
    \begin{tabular}{l c c c }
    \toprule
        & \texttt{all} & \texttt{source} & \texttt{math} \\
        \midrule
        Total premises   & \num{96915} & \num{28784} & \num{67462} \\
        Total examples   & \num{41755} & \num{20571} & \num{40187} \\
        Premises per example & 3.12  & 2.35  & 2.09 \\
    \bottomrule
    \end{tabular}
\end{table}

\begin{enumerate}
\item The \texttt{all} filter preserves almost all premises from the original, raw list, removing those that were generated automatically by Lean.
They contain a leading underscore in their names, e.g., \lstinline{RingTheory.MatrixAlgebra._auxLemma.1}.
In our example, there are no such premises.
Examples from this filter are not appropriate for training an ML advisor for interactive use as the suggestions would contain many lemmas used implicitly by tactics.
Yet, such an advisor could be used for automated ITP approaches such as \textit{hammers} \cite{hammers}.

\item The \texttt{source} filter leaves only those premises that appear in the proof's source code.
The idea is to model the explicit usage of premises by a user.
Following our example, we would take the following proof as a string and list only the three premises appearing there:
\begin{lstlisting}
by rw [div_eq_mul_inv]; exact mul_ne_zero ha (inv_ne_zero hb)
\end{lstlisting}

\item The \texttt{math} filter preserves only lemmas that are clearly of mathematical nature, discarding basic, technical ones.
The names of all theorems and definitions from \textsf{mathlib} are extracted and used as a \textit{white list}.
In particular, this means that many basic lemmas from Lean's core library (e.g. \lstinline{Eq.refl} from our example) are filtered out.
\end{enumerate}

In addition to our base datasets containing \textit{one data point per theorem}, we also created a dataset (labeled as \texttt{intermediate}) representing \textit{intermediate proof states}.
To this end, we used \textsf{LeanInk},\footnote{\url{https://github.com/leanprover/LeanInk}} a helper tool for Alectryon \cite{Alectryon+SLE2020} -- a toolkit that aids exploration of tactical proof scripts without running the proof assistant.
Given a Lean file, \textsf{LeanInk} generates all the states that a user might be able to see in the infoview (a panel in Lean that displays goal states and other information about the prover's state) by clicking on the file.
The file is split into fragments, each containing a string of Lean code, represented by a list of tokens, together with the proof states before and after.
In this way, the file can be loaded statically simulating the effect of running Lean.
Furthermore, it can be configured to keep track of typing information, which is key to detecting which tokens are premises.
We modified \textsf{LeanInk} so that every fragment that appears inside a proof is treated as its own theorem by our extractor.
We gather all the premises found in the list of tokens and featurize the hypotheses and goals in the ``before'' proof state.

This dataset consists of \num{91292} examples and \num{143165} premises, which gives an average of around $1.57$ premises per example.
It represents more fine-grained use of the premises, which is not exactly corresponding to our main objective of providing rankings of premises on the level of theorem statements.
We treat it as an auxiliary dataset potentially useful for augmenting our base data sets.


\section{Machine Learning Models}
\label{sec:ml}

The task modelled here with ML is predicting a ranking of likely useful premises (lemmas and theorems) conditioned by the features of the statement of a theorem being proved by a user.
The nature of this problem is different than common applications of classical ML: both the number of features and labels (premises) to predict is large, and the training examples are sparse in the feature space.
Thus, we could not directly rely on traditional implementations of ML algorithms, and using custom-built versions was necessary.
As one of our design requirements was tight integration with the proof assistant, we implemented the ML algorithms directly in Lean 4, without needing to call external tools.
This also served as a test for the maturity and efficiency of Lean 4 as a programming language.

In Sections \ref{sec:knn} and \ref{sec:rf} we describe two machine learning algorithms implemented in this work: $k$-nearest neighbours ($k$-NN) and random forest.

\subsection{\textit{\textbf{k}}-Nearest Neighbours}
\label{sec:knn}

This is a classical and conceptually simple ML algorithm \cite{ml}, which has already been used multiple times for premise selection \cite{premsel-isabelle,premsel-flyspeck,mizar40}.
It belongs to the \textit{lazy learning} category, meaning that it does not result in a prediction model trained beforehand on the dataset, but rather the dataset is an input to the algorithm while producing the predictions.

Given an unlabeled example, $k$-NN produces a prediction by extracting the labels of the $k$ most similar examples in the dataset and returning an averaged (or most frequent) label.
In our case, the labels are lists of premises.
We compose multiple labels into a ranking of premises according to the frequency of appearance in the concatenated labels.

The similarity measure in the feature space calculates how many features are shared between the two data points, but additionally puts more weight on those features that are rarer in the whole training dataset $\mathcal{D}$.
The formula for the similarity of the two examples $x_1$ and $x_2$ associated with sets of features $f_1$ and $f_2$, respectively, is given below.
\[
M(x_1, x_2) = \frac{
    \sum_{f \in f_1 \cap f_2} t(f)
}{
    \sum_{f \in f_1} t(f) +
    \sum_{f \in f_2} t(f) -
    \sum_{f \in f_1 \cap f_2} t(f)
}, \
t(f) = \log\left(\frac{|\mathcal{D}|}{|\mathcal{D}_f|}\right)^2,
\]
where $\mathcal{D}_f$ are those training examples that contain the feature $f$.

The advantages of $k$-NN are its simplicity and the lack of training.
A disadvantage, however, is the need to traverse the whole training dataset in order to produce a single prediction (a ranking).
This may be slow, and thus not optimal for interactive usage in proof assistants.

\subsection{Random Forest}
\label{sec:rf}

As an alternative to $k$-NN, we use \textit{random forest} \cite{rf} -- an ML algorithm from the \textit{eager learning} category, with a separate training phase resulting in a prediction model consisting of a collection of decision trees.
The leaves of the trees contain labels, and their nodes contain decision rules based on the features.
In our case, the labels are sets of premises, and the rules are simple tests that check if a given feature appears in an example.

When predicting, unlabeled examples are passed down the trees to the leaves, the reached labels are recorded, and the final prediction is averaged across the trees via voting.
The trees are trained in such a way as to avoid correlations between them, and the averaged prediction from them is of better quality than the prediction from a single tree.

Our version of random forest, adapted to deal with sparse binary features and a large number of labels, is similar to the one used in \cite{tactician}.
It is trained in an \textit{online} manner, i.e., it is updated sequentially with single training examples -- not with the entire training dataset at once, as is typically done.
The rationale for this is to make it easy to update the model with data coming from new theorems proved by a user.
This allows the model to immediately provide suggestions taking into account these recently added theorems.\footnote{This mode, however, has not yet been tested in the current stage of this work.}

\begin{SCfigure}
    \caption{\label{fig:tree}
    A schematic example of a decision tree from a trained random forest.
    Lowercase letters (\texttt{a}, \texttt{b}, \texttt{c}, ...) designate features of theorem statements, whereas uppercase letters (\texttt{P}, \texttt{Q}, \texttt{R}, ...) designate names of premises.
    The input (a featurized theorem statement) is being passed down the tree (along the green arrows) so that each node tests for a presence of a single feature, and passes the input example to the left (or right) sub-tree in the negative (or positive) case.
    The output is a set of premises in the reached leaf.
    }
    \includegraphics[width=0.48\linewidth]{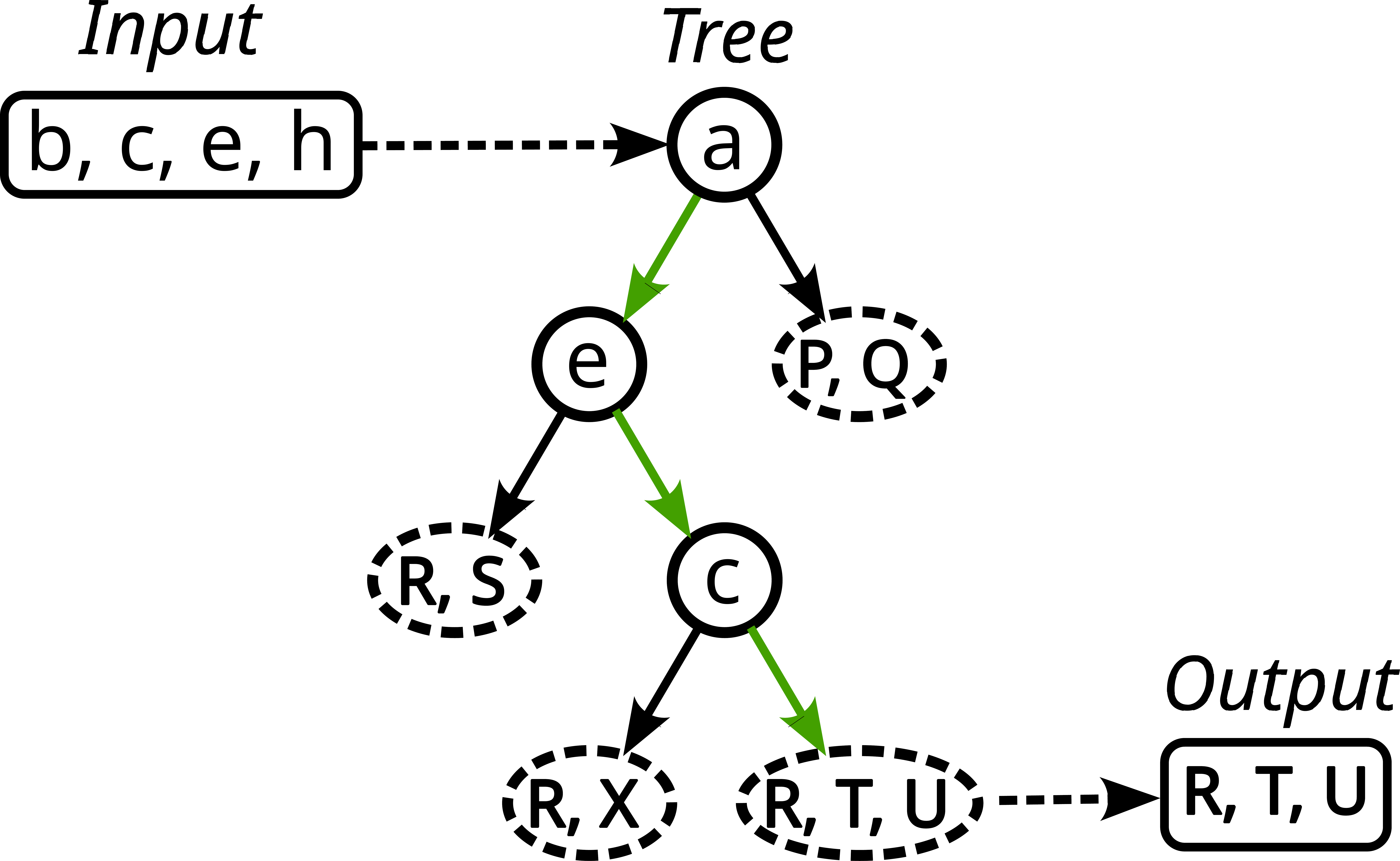}
\end{SCfigure}

Algorithm \ref{alg:tree} provides a sketch of how a training example updates a tree -- for all the details see the actual implementation in our public GitHub repository.\footnote{\url{https://github.com/BartoszPiotrowski/lean-premise-selection}}
A crucial part of the algorithm is the \textsc{MakeSplitRule} function creating node splitting rules.
Searching for the rules resulting in optimal splits would be costly, thus this function relies on heuristics.

Figure \ref{fig:tree} schematically depicts how a simple decision tree from a trained random forest predicts a set of premises for an input example.

\begin{algorithm}[h]
\algsetblockdefx[match]{Match}{EndMatch}{}{}[1]{\textbf{match} #1 \textbf{with}}{aa}
\algsetblockdefx[with]{With}{EndWith}{}{}[1]{{#1}\textbf{:}}{aa}
\algnotext[with]{EndWith}
\algnotext[match]{EndMatch}
\caption{Updating a tree with a training example in a random forest.}
\label{alg:tree}
\begin{algorithmic}[1]
\Function {AddExampleToTree}{$T$, $e$} \Comment{$T$ -- tree to update, $e$ -- training example}
\Match{$T$}
    \With{Node($R$, $T_l$, $T_r$)}
    \Comment{$R$ -- binary rule, $T_l$, $T_r$ -- left and right subtrees}
        \Match{$R$($e$)}
        \Comment{passing example $e$ down the tree to a leaf}
            \With{Left}
            \Return Node($R$, \textsc{AddExampleToTree}($T_l$, $e$), $T_r$)
            \EndWith
            \With{Right}
            \Return Node($R$, $T_l$, \textsc{AddExampleToTree}($T_r$, $e$))
            \EndWith
        \EndMatch
    \EndWith
    \With{Leaf($E$)}
    \Comment{$E$ -- examples stored in the leaf}
        \State $E$ $\gets$ \textsc{Append}($E$, $e$)
        \If {\textsc{SplitCondition}($E$)}
        \Comment{testing if the leaf should be split}
            \State $R$ $\gets$ \textsc{MakeSplitRule}($E$)
            \Comment{making semi-optimized new split rule}
            \State $E$$_l$, $E$$_r$ $\gets$ \textsc{Split}($R$, $E$)
            \Comment{splitting examples into two parts}
            \State \Return Node($R$, Leaf($E_l$), Leaf($E_r$))
            \Comment{new subtree growing the tree}
        \Else
            \State \Return Leaf($E$)
            \Comment{the original leaf augmented with example $e$}
        \EndIf
    \EndWith
\EndMatch
\EndFunction
\end{algorithmic}
\end{algorithm}

\section{Evaluation Setup and Results}
\label{sec:results}

To assess the performance of the ML algorithms, the data points extracted from \textsf{mathlib} were split into \textit{training} and \textit{testing} sets.
The testing examples come from the modules that are \textit{not} dependencies of any other modules (there are 592 of them).
This simulates a realistic scenario in which a user utilizing the suggestion tool develops a new \textsf{mathlib} module.
The rest of the modules (2436) served as the source of training examples.

Two measures of the quality of the rankings produced by ML are defined: Cover and Cover$_+$.
Assuming a theorem $T$ depends on the set of premises $P$ of size $n$, and $R$ is the ranking of premises predicted by the ML advisor for $T$, these measures are defined as follows:
\[
\text{Cover}(T) = \frac{\big|P \cap R\texttt{[:} n\texttt{]}\big|}{n},
\hspace{1cm}
\text{Cover$_+$}(T) = \frac{\big|P \cap R\texttt{[:} n+10\texttt{]}\big|}{n},
\]
where $R\texttt{[:} k\texttt{]}$ is a set of $k$ initial premises from ranking $R$.
Both Cover and Cover$_+$ return values in $[0,1]$.
Cover gives the score of $1$ only for a ``perfect'' prediction where the premises actually used in the proof form an initial segment of the ranking.
Cover$_+$ may also give a perfect score to less precise predictions.
The rationale for Cover$_+$ is that the user in practice may look through 10 or more suggested premises.
This is often more than the $n$ premises actually used in the proof, so we consider initial segments of length $n + 10$ in Cover$_+$.

Both $k$-NN and random forest are evaluated on data subject to all three premise filters described in Section \ref{sec:filters}.
For each of these variants of data, three combinations of features are tested: (1) \texttt{names} only, (2) \texttt{names} and \texttt{bigrams}, (3) \texttt{names}, \texttt{bigrams}, and \texttt{trigrams}.
The hyper-parameters for the ML algorithms were selected by an experiment on a smaller dataset.
For $k$-NN, the number of neighbours was fixed to $100$.
For random forest, the number of trees was set to $300$, each example was used for training a particular decision tree with probability equal to $0.3$, and the training algorithm passed through the whole training data $3$ times.

Table \ref{tab:results} shows the results of the experiment.
In terms of the Cover metric, random forest performed better than $k$-NN for all data configurations.
However, for Cover$_+$ metric, $k$-NN surpassed random forest for the \texttt{math} filter.

It turned out that the union of \texttt{names} and \texttt{bigrams} constitutes the best features for all the filters and both ML algorithms.
It likely means that the more complex \texttt{trigrams} did not help the algorithms to generalize well but rather caused \textit{over-fitting} on the training set.

The results for the \texttt{all} filter appear to be much higher than for the other two filters.
However, this is because applying \texttt{all} results in many simple examples containing just a few common, basic premises (e.g., just a single \texttt{rfl} lemma).
They increase the average score.

Overall, random forest with \texttt{names}$\, + \,$\texttt{bigrams} (\texttt{n+b}) features gives the best results.
An additional practical advantage of this model over $k$-NN is the speed of outputting predictions.
For instance, for the \texttt{source} filter and \texttt{n+b} features, the average times of predicting a ranking of premises per theorem were \SI{0.28}{\second} and \SI{5.65}{\second} for random forest and $k$-NN, respectively.

Additionally, we evaluated the ML models on the \texttt{intermediate} dataset, using \texttt{n+b} features.
The random forest achieved Cover = 0.09 and Cover$_+$ = 0.24, whereas $k$-NN resulted in Cover = 0.08 and Cover$_+$ = 0.21 on the testing part of the data.
Then, we used the \texttt{intermediate} dataset in an attempt to improve the testing results on the base dataset with the \texttt{source} filter (as \texttt{intermediate} only contains premises visible in the source files).
We used the \texttt{intermediate} data as a \textit{pre-training} dataset,
first training a random forest on it, and later on the base data.
We also used \texttt{intermediate} to \textit{augment} the base data, mixing the two together.
However, neither in the pre-training, nor in the augmentation mode statistically significant improvements in the testing performance were achieved.
It is possible that the prediction quality from the practical perspective actually improved, being more proof-state-dependent and not only theorem-dependent, but it did not manifest in our theorem-dependent evaluation.

The evaluation may be reproduced by following the instructions in the linked source code.\footnote{\url{https://github.com/BartoszPiotrowski/lean-premise-selection\#reproducing-evaluation}}

\begin{table}[t]
\caption{\label{tab:results}
Average performance of random forest and $k$-NN on testing data, for three premises filters and three kinds of features.
The type of features is indicated by a one-letter abbreviation: \texttt{n} = \texttt{names}, \texttt{b} = \texttt{bigrams}, \texttt{t} = \texttt{trigrams}.
For each configuration, Cover and Cover+ measures are reported (the latter in brackets).
In each row, the best Cover result is bolded.
}
\centering
\begin{tabular}{l@{\hspace{0.2cm}}ccc@{\hspace{0.3cm}}ccc@{\hspace{0.2cm}}}
\toprule
	&\multicolumn{6}{c}{\textbf{machine learning model}} \\
    \cmidrule[0.1em](r){2-7}
    &\multicolumn{3}{c}{random forest} & \multicolumn{3}{c}{$k$-nearest neighbours} \\
	\cmidrule[0.06em](r){2-4} \cmidrule[0.06em](r){5-7}
\thead{\textbf{premises}} & \texttt{n} & \texttt{n+b} & \texttt{n+b+t} & \texttt{n} &
\texttt{n+b} & \texttt{n+b+t} \\
\cmidrule[0.05em](r){1-1}
\cmidrule[0.05em](r){2-4}
\cmidrule[0.05em](r){5-7}
\texttt{all}     & \thead{0.56 (0.67)} & \thead{\textbf{0.57} (0.67)} & \thead{0.47 (0.58)} & \thead{0.51 (0.65)} & \thead{0.52 (0.66)} & \thead{0.51 (0.62)} \\
\texttt{source}  & \thead{0.28 (0.36)} & \thead{\textbf{0.29} (0.36)} & \thead{0.28 (0.36)} & \thead{0.25 (0.35)} & \thead{0.25 (0.36)} & \thead{0.26 (0.35)} \\
\texttt{math}    & \thead{0.25 (0.32)} & \thead{\textbf{0.26} (0.33)} & \thead{0.16 (0.24)} & \thead{0.22 (0.34)} & \thead{0.23 (0.34)} & \thead{0.16 (0.26)} \\
\bottomrule
\end{tabular}
\end{table}

\section{Interactive Tool}
\label{sec:tool}

\begin{figure}
    {\centering
    \hspace{-16pt}
    \includegraphics[width=1.07\linewidth]{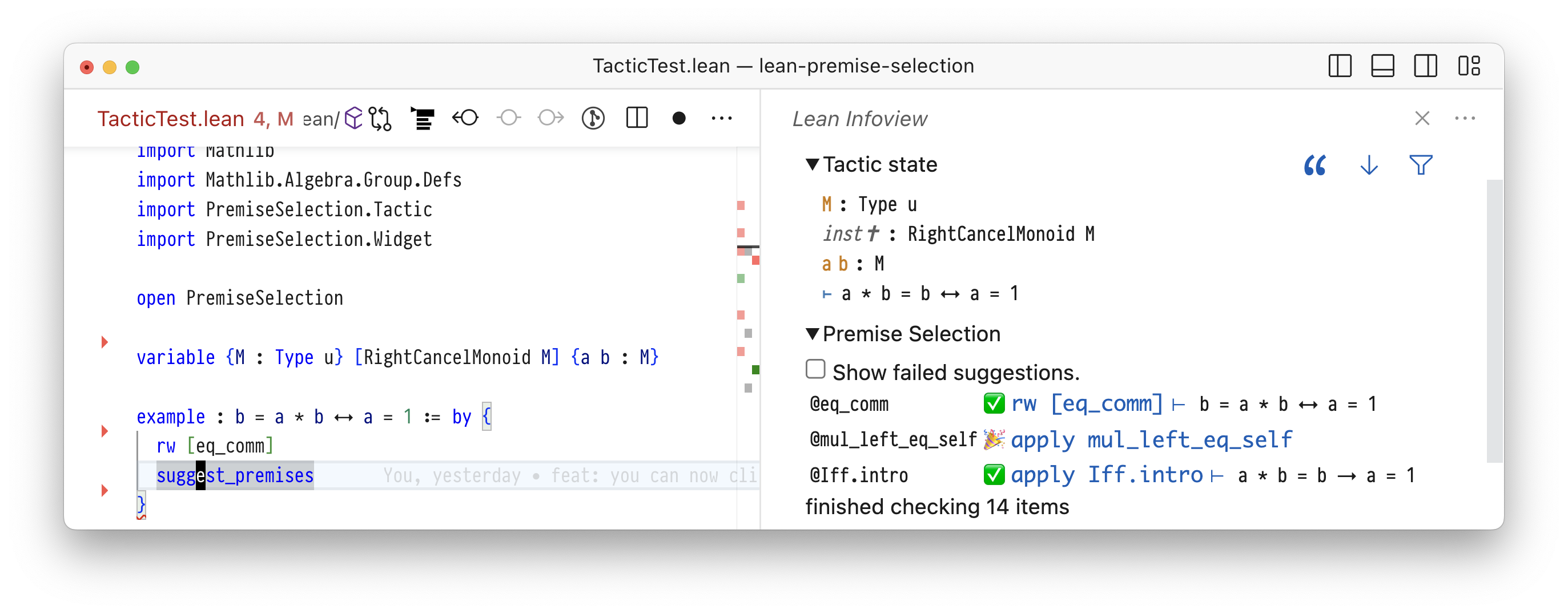}}
    \vspace{-20pt}
    \caption{\label{fig:ui}
    The interactive tool in Visual Studio Code. The left pane shows the source file with the cursor over a \lstinline{suggest_premises} tactic. The right pane shows the goal state at the cursor position and, below, the suggested lemmas to solve the goal.
    Suggestions annotated with a checkbox advance the goal state, suggestions annotated with confetti close the current goal. Clicking on a suggested tactic (e.g. \lstinline{apply mul_left_eq_self}) automatically appends to the proof script on the left.}
\end{figure}

The ML predictor is wrapped in an interactive tactic \lstinline{suggest_premises} that users can type into their proof script.
It will invoke the predictor and produce a list of suggestions.
This list is displayed in the infoview.
The display makes use of the new remote-procedure-call (RPC) feature in Lean 4 \cite{widgets}, to then asynchronously run various tactics for each suggestion.
Given a suggested premise $p$, the system will attempt to run tactics \lstinline{apply }$p$, \lstinline{rw [}$p$\lstinline{]} and \lstinline{simp only [}$p$\lstinline{]}, and return the first successful tactic application that advances the state. This will then be displayed to the user as shown in Figure \ref{fig:ui}. She can select the resulting tactic to insert into the proof script.
By using an asynchronous approach, we can display results rapidly without waiting for a slow tactic search to complete.

\section{Future Work}
\label{sec:future}

There are several important directions in which the current work may be developed.
The results may be improved by augmenting the dataset with, for instance, synthetic theorems, as well as developing better features, utilizing the well-defined structure of Lean expressions.

Applying modern neural architectures in place of the simpler ML algorithms used here is a promising path \cite{premsel-aitp}.
It would depart from our philosophy of a lightweight, self-contained approach as the suggestions would come from an external tool, possibly placed on a remote server.
However, given the strength of the current neural networks, we could hope for higher-quality predictions.
Moreover, neural models do not require hand-engineered features.

Finally, premise selection is an important component of ITP \textit{hammer} systems \cite{hammers}.
The presented tool may be readily used for a hammer in Lean, which has not yet been developed.

\bibliographystyle{splncs04}
\bibliography{bibliography}
\end{document}